\documentclass[a4paper]{article}

\usepackage{pdfpages}

\usepackage[english]{babel}
\usepackage[utf8x]{inputenc}
\usepackage[T1]{fontenc}
\usepackage{amsthm}

\usepackage[a4paper,top=3cm,bottom=2cm,left=3cm,right=3cm,marginparwidth=1.75cm]{geometry}

\usepackage{booktabs}
\usepackage{amsmath}
\usepackage{amssymb}
\usepackage{graphicx}
\usepackage[colorlinks=true, allcolors=blue]{hyperref}
\usepackage{url}
\usepackage{caption}
\usepackage{subcaption}
\DeclareCaptionType{Figure}

\theoremstyle{definition}  
\newtheorem{theorem}{Theorem}
\newtheorem{proposition}{Proposition}

\newtheorem*{theorem*}{Theorem}

\newcommand{\sitessigma}[0]{\text{sites}\; \sigma}
\newcommand{\asvector}[1]{{#1}}
\newcommand{\tangentspace}[1]{\mathrm{T}_{\!#1}}
\newcommand{\gradient}[2]{\nabla_{\!#1}\,#2}

\newcommand{\reals}[0]{\mathbb{R}}

\newcommand{\pointconfig}[1]{\mathbf #1}
\newcommand{\pcpoint}[2]{{#1}^{#2}}
\newcommand{\pointi}[0]{\pcpoint{x}{i}}
\newcommand{\pointj}[0]{\pcpoint{x}{j}}

\newcommand{\hyperboloid}[2]{\mathbb{H}_{#1}^{#2}}
\newcommand{\pb}[2]{\mathbb{B}_{#1}^{#2}}  
\newcommand{\minkowski}[1]{\mathbb{R}^{#1:1}}

\newcommand{\minkowskiformpower}[4]{\left\langle {#2},{#3} \right\rangle^{#4}}
\newcommand{\minkowskiform}[3]{\minkowskiformpower {#1} {#2} {#3} {}}

\newcommand{\dist}[0]{\mathrm{d}}

\newcommand{\likelihood}[0]{\mathcal{L}}
\newcommand{\objective}[0]{\mathbf{l}}
\newcommand{\probamatrix}[0]{P}
\newcommand{\proba}[1]{{\probamatrix}_{\!#1}}
\newcommand{\probadiag}[0]{\proba {\scriptscriptstyle \text{diag}}}
\newcommand{\probaoffdiag}[0]{\proba {\scriptscriptstyle \lnot \text{diag}}}
\newcommand{\sigmai}[0]{\sigma_{\!i}}
\newcommand{\sigmaj}[0]{\sigma_{\!j}}
\newcommand{\distfnsinglevar}[1]{\dist_{#1}}
\newcommand{\pcwithouti}[0]{{\pointconfig x \setminus \pointi}}

\DeclareMathOperator{\Exp}{Exp}
\DeclareMathOperator{\Log}{Log}

\DeclareMathOperator{\arccosh}{arccosh}

\title{Learning phylogenetic trees as hyperbolic point configurations}

\author{
  Benjamin Wilson\\
	Lateral GmbH\\
  \texttt{benjamin@lateral.io}
}
\date{
\today
}

\begin{document}
\maketitle

\begin{abstract}
We propose a novel method for the inference of phylogenetic trees that utilises
point configurations on hyperbolic space as its optimisation landscape.  Each
taxon corresponds to a point of the point configuration, while the evolutionary
distance between taxa is represented by the geodesic distance between their
corresponding points.  The point configuration is iteratively modified to
increase an objective function that additively combines pairwise log-likelihood
terms.  After convergence, the final tree is derived from the inter-point
distances using a standard distance-based method.  The objective function,
which is shown to mimic the log-likelihood on tree space, is a differentiable
function on a Riemannian manifold. Thus gradient-based optimisation techniques
can be applied, avoiding the need for combinatorial rearrangements of tree
topology.
\end{abstract}

\begin{section}{Introduction}
We propose a method that iteratively rearranges the points representing the taxa to
maximise an objective function that mimics the (log-)likelihood function on
tree space.
The novelty of the proposal lies in its use of a parameter space that is
entirely smooth, consisting of the positions of points on a Riemannian
manifold.
Thus, by choosing an objective function that is smooth and does not depend upon
a choice of tree topology (the pairwise likelihood of Holder and Steel
\cite{HolderSteel2011}), gradient-based optimisation techniques can be applied
consistently throughout the process of tree inference.
This contrasts with the approach of maximum likelihood tree search.  
Tree search alternates between continuous optimisation of the branch lengths and discrete optimisation
of the tree topology.  The discrete optimisation step entails computing the
best likelihood of each tree topology that is neighbouring under combinatorial
tree re-arrangement operations, such as nearest neighbour interchange (NNI) or
subtree prune and regraft (SPR); the best neighbour is then adopted as the new
tree topology \cite{MathEvoPhyl}.  The parameter space of maximum likelihood
tree search has therefore both continuous and discrete (graph-like) structure.
Inherent in such methods are design choices about which combinatorial tree
re-arrangements define the combinatorial landscape.  By contrast,
in the proposed method, the nearness of trees arises directly from the
underlying hyperbolic geometry.

The paper is structured as follows.  Section \ref{Section:general-approach}
illustrates and motivates the approach.  Section
\ref{Section:four-point-condition} shows how hyperbolic space satisfies a
weakened version of the four-point condition (4PC), while supporting background
on hyperbolic geometry is contained in the appendices.  Section
\ref{Section:curvature-and-dimension} illustrates how the curvature and
dimension of hyperbolic space influence its ability to approximate tree
metrics.  Section \ref{Section:learning} reviews the adaption of the pairwise
likelihood to our optimisation landscape,
 illustrates the relationship of the objective function to
the likelihood on trees, and describes how it can be maximised numerically
using gradient ascent.  Section \ref{Section:implementation} discusses the
implementation, Section \ref{Section:experiments} compares the performance of
the proposed method to
that of established tree inference methods, and Section
\ref{Section:discussion} discusses possible directions for future research.
\end{section}

\begin{section}{Acknowledgments}
I am profoundly grateful to Lateral GmbH for material
support of this project.  I thank Matthias Leimeister, Stephen Enright-Ward,
Ben Kaehler, Jeremy Sumner, Venta Terauds and the participants
of the Phylomania conferences for their interest and helpful discussions.
\end{section}

\begin{section}{Illustration of approach}\label{Section:general-approach}
Let $M$ be a Riemannian manifold (e.g. $M=\reals^m$) with distance function $\dist_M$, and let $N > 0$.
A {\it point configuration on $M$ of size $N$} is an ordered collection $\pointconfig x = (\pointi)_{1 \leqslant i \leqslant N}$ of points $\pointi \in M$.
Equivalently, a point configuration of size $N$ is an element of the product manifold $M^N$.
For clarity, we may refer to the pairwise distances between the points of a point configuration as {\it inter-point distances}.

Suppose now that there are $N$ taxa, labelled by the indices $1, \dots, N$, and let $\Theta$ denote homologous sequence data of length $L$.
Write $\likelihood_{ij} (t | \Theta)$ for the likelihood of an evolutionary distance of $t$ between taxa $i$, $j$ given the sequence data $\Theta$.
Define the function 
\begin{equation}\label{logalike-introduction}
	\objective : M^N \rightarrow \reals, \quad
	\objective : \pointconfig x \mapsto
	\tfrac{1}{L} \sum_{i \neq j}
	\log \likelihood_{ij} (\dist_M (\pointi, \pointj) | \Theta),
\end{equation}
for any point configuration $\pointconfig x \in M^N$.
The function $\objective$ is the scaled sum of the log-likelihoods of each inter-taxa distance being given by the corresponding inter-point distance.  It is the "pairwise likelihood" introduced in \cite{HolderSteel2011}, where it was considered as a function of the inter-leaf distances on a tree.  We consider it as a function of the inter-point distances of a point configuration $\pointconfig x$, and seek point configurations that maximise its value (it is the {\it objective function} of our optimisation).

The function $\objective$ incorporates, at least formally, some important aspects of the log-likelihood on tree space.
Unlike the log-likelihood on tree space, however, its argument is a point on a Riemannian manifold (i.e. $\pointconfig x \in M^N$), and it is moreover differentiable in this argument.
Thus, just as in the case when $M = \reals^m$, hill climbing techniques from multivariate calculus can be applied to find a local maximum of the function $\objective$.
For example, a point configuration $\pointconfig x \in M^N$ can be modified to another point configuration with higher $\objective$ value by, for each point $\pointi$, considering $\objective$ as a function of that point only, and moving the point a small distance in the direction of the gradient.
Maximising $\objective$ through many of these small updates, it might be hoped (given the resemblance of $\objective$ to the log-likelihood) that the inter-point distances of $\pointconfig x$ come to approximate the inter-leaf distances of a tree that produced, with high likelihood, the sequence data $\Theta$.
Indeed, more elaborate optimisation approaches, for example involving second-order derivatives, may equally be applied to maximise $\objective$ \cite{boumal2020intromanifolds}.

The inter-point distances of a point configuration $\pointconfig x$ are interdependent.  That is, it is not possible to change $\dist_M (\pointi, \pointj)$ without moving $\pointi$ (or $\pointj$), and doing so will, in general, change also the other inter-point distances involving the moved point.
Thus the maximisation just discussed effects a {\it joint} estimation of the inter-leaf distances.
The manner in which the distances are interdependent is controlled by the choice of manifold $M$.
This paper is concerned with the choice of a Riemannian manifold $M$ and the design and optimisation of functions $\objective$ for the purpose of approximating the inter-leaf metric of a tree that, with high likelihood, generated the given sequence data $\Theta$.
If the approximation is sufficiently good, then this high-likelihood-tree can be recovered from the inter-leaf distances via e.g. the neighbour joining algorithm.

\begin{figure}
	\centering
	\includegraphics[width=\textwidth]{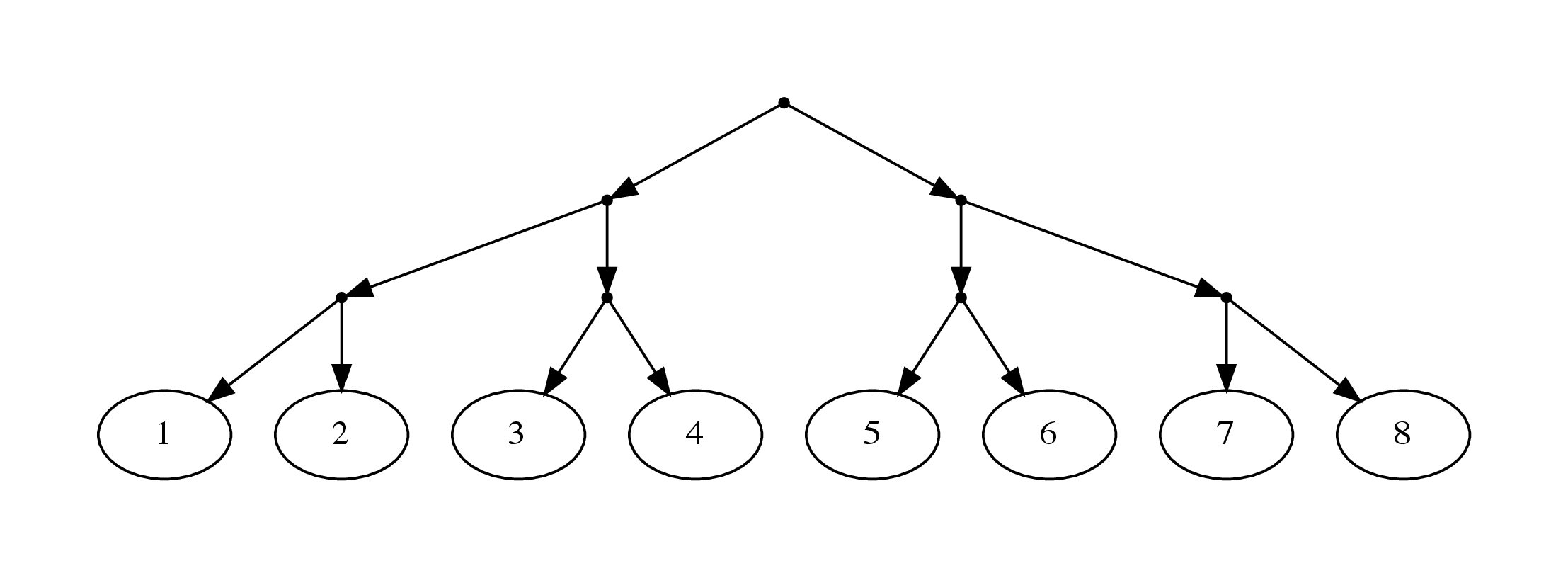}
	\caption{A balanced binary tree with $8$ leaves.}
	\label{balanced-binary-tree}
\end{figure}

\begin{figure}
	\centering
	\includegraphics[width=0.6\textwidth]{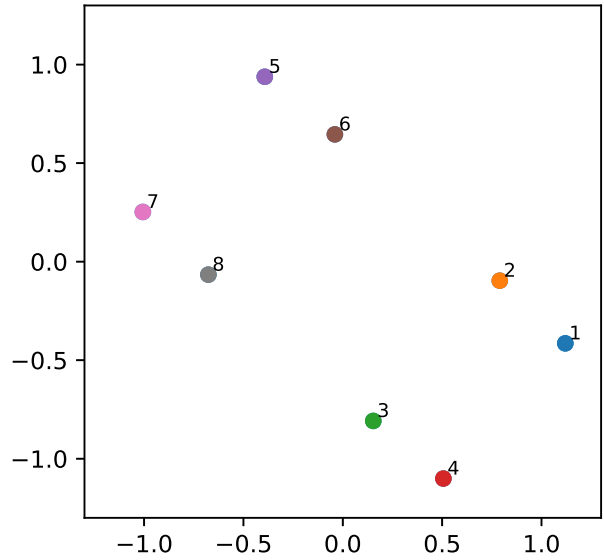}
	\caption{A point configuration on $M=\reals^2$ maximising the objective $\objective$, given sequence data generated by the balanced binary tree with $8$ leaves and all edge lengths equal to $\epsilon=0.25$ (cf. Figure \ref{balanced-binary-tree}}
	\label{euclidean-fit-1-article}
\end{figure}

Consider now the case of a balanced binary tree (cf. Figure \ref{balanced-binary-tree}) with $N = 2^k$ leaves, and all
edge lengths\footnote{Here and throughout, edge lengths indicate the amount of
evolutionary time elapsed.} equal to $\epsilon$.
Figure \ref{euclidean-fit-1-article} depicts a point configuration on $M=\reals^2$
chosen to maximise the function $\objective$, given sequence data of length
$L=800$.
The tree has a diameter of $k \epsilon$.
Thus the points of a point configuration maximising $\objective$ can not be too far from
one another, and indeed are contained in a ball of radius proportional to $k
\epsilon$.  On the other hand, the volume of such a ball is only quadratic in
the diameter $k \epsilon$, yet must accommodate a number of leaves $N=2^k$ that
is exponential in $k$.  Thus the Euclidean plane suffers from an "overcrowding"
problem for large values of $k$, and so is a poor choice of manifold $M$.
Indeed, this problem persists in higher dimensional Euclidean space
$M=\reals^m$, since the volume of a ball, being a degree $m$ polynomial in the
diameter, is still outgrown by the exponential growth in the number of leaves
when $k$ is sufficiently large.  On the other hand, hyperbolic space is a
Riemannian manifold on which the volume of a ball is {\it exponential} in the
diameter.  Thus hyperbolic space does not suffer from this overcrowding
problem, and is a good choice for the manifold $M$.  The advantage of
hyperbolic space for our purpose is treated formally in Section
\ref{Section:four-point-condition} in terms of the $\delta$-hyperbolicity and
the 4PC, and further illustrated in Section
\ref{Section:curvature-and-dimension} where the influence of curvature is
demonstrated.
\end{section}
\begin{section}{Hyperbolic space and the four point condition}\label{Section:four-point-condition}
This section demonstrates the suitability of hyperbolic space as a setting for the proposed approach.
It is shown that hyperbolic space (but not Euclidean space) satisfies a
weakened version of the 4PC, and that the maximal
violation of the 4PC is a linear function of the radius $\rho$ of the
hyperboloid.  Thus the 4PC can be approximated to any desired accuracy by
taking $\rho$ sufficiently small.

Metrics arising as the inter-leaf distances on trees are characterised by the 4PC:

\begin{theorem}\label{Theorem:Buneman}\cite{Zaretskii1965,Buneman1971}
Let $D=(\dist(x, y))_{x,y \in X}$ be the matrix of pairwise distances of a finite metric space $(X, \dist)$.
Then there exists a tree with inter-leaf distances given by $D$ if and only if
	\begin{equation}\label{four-point-condition}
	\dist(x, w) + \dist(y, z) \leqslant \max \{ \dist(x, y) + \dist(z, w),\ \dist(x, z) + \dist(y, w) \}
	\end{equation}
for all $w, x, y, z \in X$.
\end{theorem}

The $\delta$-hyperbolicity of Gromov\cite{Alonsoetal} can be viewed as a relaxation of the four
point condition.  Let $\delta \geqslant 0$.  A metric space $(X, \dist)$ is
said to be $\delta$-hyperbolic\footnote{There are various definitions of
$\delta$-hyperbolicity.  We note that this definition is only equivalent to the $\delta$-hyperbolicity definition in terms of 
"slim triangles" up to the choice of a different value of $\delta$,
c.f. \cite{Alonsoetal}.} if, for all $w, x, y, z \in X$,
\begin{equation}\label{hyperbolicity-equation}
\dist(x, w) + \dist(y, z) \leqslant \max \{ \dist(x, y) + \dist(z, w),\ \dist(x, z) + \dist(y, w) \} + 2 \delta.
\end{equation}
Thus a metric space is $\delta$-hyperbolic if the error of the four point
condition is bounded by $2\delta$.  By Theorem \ref{Theorem:Buneman}, the
metric spaces arising from the inter-leaf distances on trees are $0$-hyperbolic,
and indeed these are all the finite $0$-hyperbolic metric spaces, up to
isometry.  On the other hand, Euclidean space $\reals^m$, for $m \geqslant 2$
is not $\delta$-hyperbolic for any $\delta \geqslant 0$.  To see this, it
suffices to consider the four point metric spaces given by the corners of
squares of ever increasing side length.  However, as we see below, hyperbolic
space {\it is} $\delta$-hyperbolic. Moreover, the minimal $\delta$ such that
$\hyperboloid \rho m$ is $\delta$-hyperbolic scales linearly with the radius
$\rho$ of the hyperboloid.

\begin{theorem}\label{Theorem:hyperbolicity-hyperbolic}
\begin{enumerate}
\item For any $m \geqslant 2$ and $\rho > 0$, there exists $\delta > 0$ such that
$\hyperboloid \rho m$ is $\delta$-hyperbolic.
\item For any $m \geqslant 2$ and $\rho > 0$, write $\delta_{m,\rho}$ for the minimal $\delta$ such that $\hyperboloid \rho m$ is $\delta$-hyperbolic.
Then $\delta_{m,\rho} = \rho \delta_{m,1}$.
\end{enumerate}
\end{theorem}
It is shown in \cite{BogdanHyperbolicity} that  $\delta_{2, 1} = \ln (2)$.
\end{section}

\begin{section}{Curvature and dimension}\label{Section:curvature-and-dimension}
This section illustrates the consequences of the discussion of Section \ref{Section:four-point-condition}, namely that, in the absence of sample noise, the inter-leaf distances of the generating tree can be approximated to an arbitrary accuracy by taking the hyperboloid radius $\rho$ sufficiently small.

Consider again the example of Section \ref{Section:general-approach} of a balanced binary tree with $N=8$ leaves (depicted in Figure \ref{balanced-binary-tree}) and where all edges have length $0.25$, taking $M = \hyperboloid 2 1$ as the chosen manifold, instead of the Euclidean plane.
Figure \ref{point-configuration} shows a point configuration on $\hyperboloid 2 1$ maximising the function $\objective$ of Section \ref{Section:general-approach} depicted on the Poincaré disc\footnote{The sequence data is taken here to be of sufficient length that sample noise is negligible.}.

The distances of the each of the eight leaves of the generating tree from leaf $1$ are given by
\begin{equation}\label{pairwise-distances}
0, 0.5, 1, 1, 1.5, 1.5, 1.5, 1.5,
\end{equation}
(respectively).  In contrast, on Figure \ref{point-configuration}, notice that the point corresponding to
leaf $1$ is not equidistant from the points corresponding to leaves $5-8$,
although these leaves are equidistant on the tree.

\begin{figure}[htb]%
	\begin{minipage}[t]{0.3\textwidth}
	\centering
	\includegraphics[width=\textwidth]{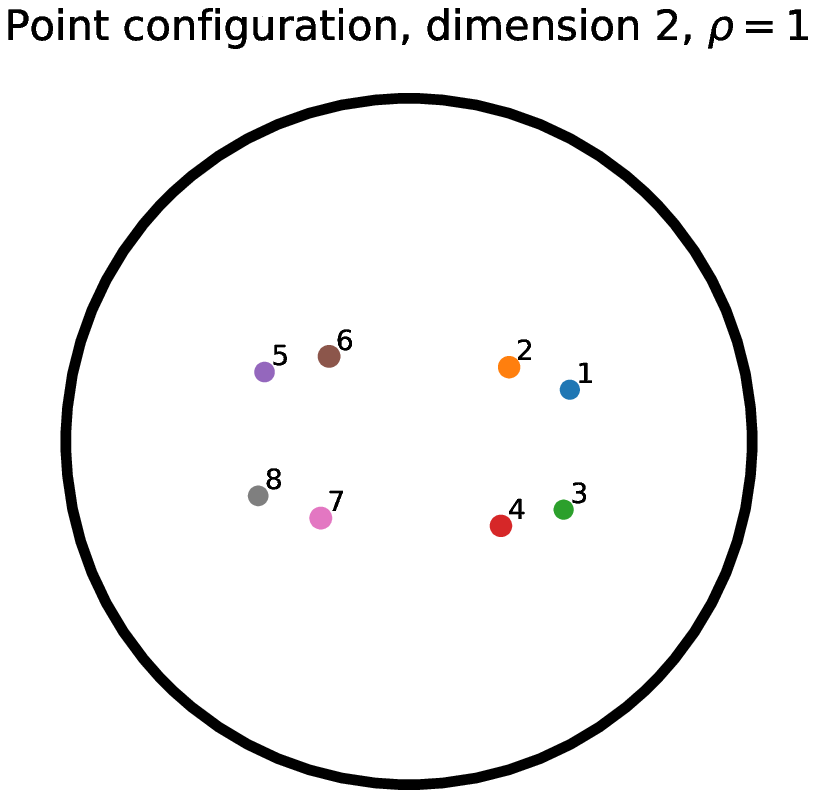}
	\subcaption{}\label{point-configuration}
	\end{minipage}%
	\hfill
	\begin{minipage}[t]{0.3\textwidth}
	\centering
	\includegraphics[width=\textwidth]{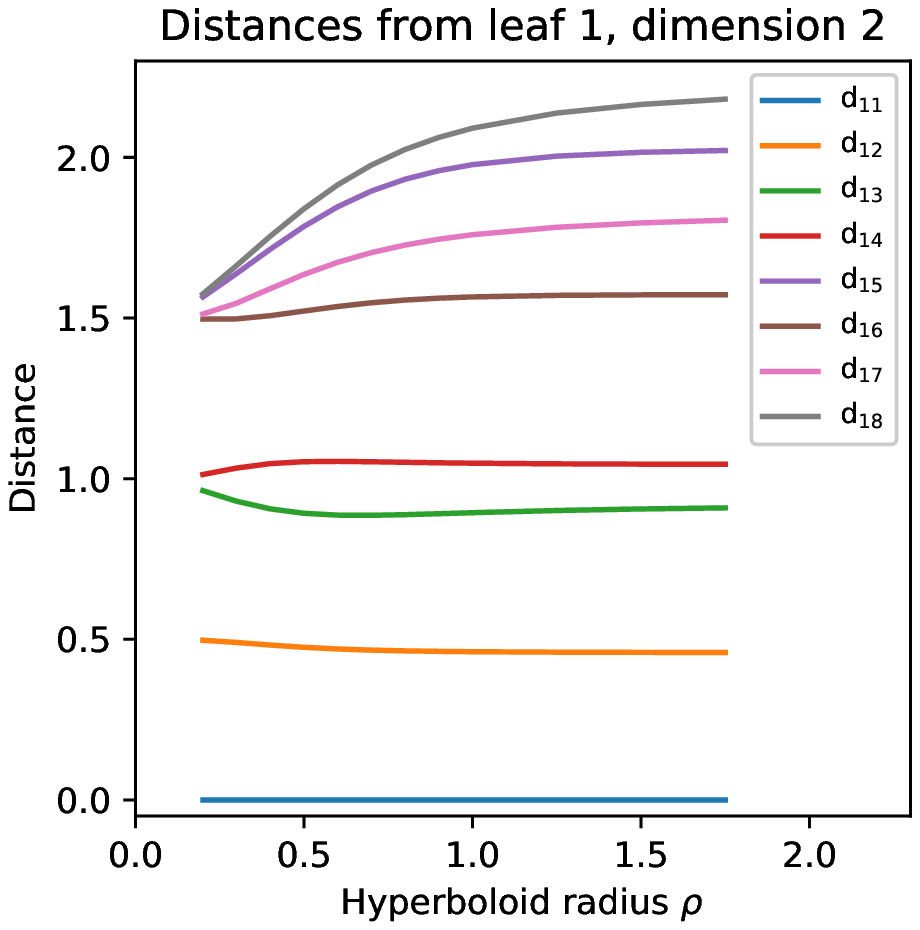}
	\subcaption{}\label{distance-rho-dependence-dimension-2}
	\end{minipage}%
	\hfill
	\begin{minipage}[t]{0.3\textwidth}
	\centering
	\includegraphics[width=\textwidth]{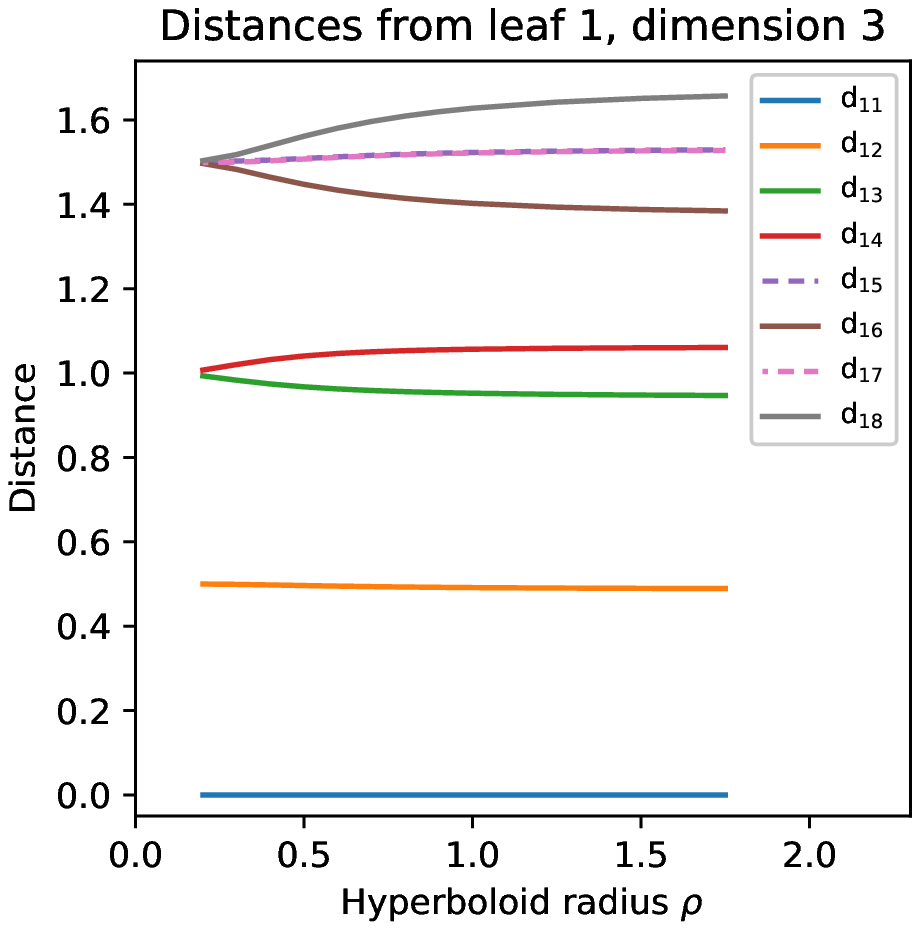}
	\subcaption{}\label{distance-rho-dependence-dimension-3}
	\end{minipage}%
	\caption{An illustration of the role of curvature (determined by the hyperboloid radius $\rho$) and dimension in the accuracy of the approximation of inter-leaf distances.
	Figure \ref{point-configuration} shows a point configuration fit to a balanced binary tree of Figure \ref{balanced-binary-tree} with $N=8$ leaves when $\rho=1$ and the dimension is $2$.
	Figure \ref{distance-rho-dependence-dimension-2} plots the pairwise distances $\mathrm{d}_{1j}$ between the point of leaf $1$ and the points representing each leaf $j$ as a function of $\rho$ in dimension $2$.
	Figure \ref{distance-rho-dependence-dimension-3} plots the same, but in dimension $3$.}
	\label{distance-rho-dependence}
\end{figure}

Indeed, the spatial distances between the points only approximate the
tree distances \eqref{pairwise-distances}.  However, this approximation
can be improved by increasing the negative curvature of the space (by decreasing the radius $\rho$ of the hyperboloid).
This is illustrated in Figure 
\ref{distance-rho-dependence-dimension-2}, which shows the spatial
distances from point $1$ to all eight points as $\rho$ is varied.  It
can be seen that lower values of $\rho$ yield approximations that
approach the true tree distances \eqref{pairwise-distances}.  On the
other hand, as $\rho$ grows larger, the space becomes more Euclidean
(i.e. flatter), and it is apparent that the distance approximations
become worse.  This demonstrates, in particular, that hyperbolic space
is a setting more appropriate to representing trees than Euclidean
space, and that the higher the negative curvature, the more
appropriate.  
The approximation is further improved by increasing the dimension from
$2$ to $3$.  Figure 
\ref{distance-rho-dependence-dimension-3} demonstrates that already in
dimension $3$, when $\rho=0.2$, the tree distances
\eqref{pairwise-distances} match the spatial distances almost
perfectly.
\end{section}

\begin{section}{Learning point configurations}\label{Section:learning}
This section treats the problem of learning a point configuration in hyperbolic
space from homologous sequence data such that the geodesic distance between the
points approximates the evolutionary distance between the taxa.  As discussed
previously, the geometry of hyperbolic space enforces a weakened four point
condition on the matrix of pairwise distances.  The points representing the
taxa are initially placed on the hyperboloid according to an embedding of the
tree obtain using Weighbor \cite{Weighbor}. 
The points are then adjusted iteratively to maximise the objective $\objective$,
which is designed to mimic the likelihood function on tree space.

	\begin{subsection}{The mutation model}\label{Subsection:mutation-model}
	Denote by $N$ the number of taxa, and for given homologous sequence data of length $L$,
	denote by $\sigmai$ the base at site $\sigma$ for taxon $i$.  Let $\probamatrix (t),
	t \geqslant 0$ be row-stochastic matrices defining a time-homogeneous, ergodic,
	stationary and time-reversible model of mutation.
	Thus the mutation model is some instance of the generalised
	time-reversible model \cite{TavareGTR}, which includes the Jukes-Cantor model as a special case.
	Write $Q = \frac{\partial}{\partial t} |_{t=0} \probamatrix (t)$ for the infinitesimal
	generator of $\probamatrix$, and $\pi$ for its stationary distribution.
	For any bases $a,b$, the entry $\proba {ab} (t)$ is the conditional probability
	of observing $b$ at any site $t$ time units after observing an $a$ there.

	Given two taxa $i$ and $j$, recall that the likelihood of an evolutionary
	distance of $t$, given their homologous sequence data, is given by
	$$\likelihood (t) = \prod_{\sitessigma} \pi_{\sigmai} \proba {\sigmai \sigmaj}(t),$$
	and so the log-likelihood has the form
	\begin{equation}\label{loglikelihood}
	\log \likelihood (t) = \sum_{\sitessigma} \log \proba {\sigmai \sigmaj}(t) + C,
	\end{equation}
	where $C = \sum_\sigma \log \pi_{\sigmai}$ is a constant that may be ignored
	when choosing $t$ to maximise $\log \likelihood (t)$ (or equivalently,
	$\likelihood (t)$).
	It follows immediately that, for any point configuration $\pointconfig x \in M^N$,
	$$
	\objective (\pointconfig x) =
	\tfrac{1}{L} \sum_{i \neq j}
	\sum_{\sitessigma} 
	\log \proba {{\sigmai} {\sigmaj}} \left(\dist (\pointi, \pointj) \right).
	$$
	up to an additive constant.  We henceforth adopt this simpler form of the objective $\objective$.

	The Jukes-Cantor model is a particularly simple mutation model, in that it
	has no parameters of its own\footnote{Assuming that the substitution rate has
	been fixed, as it is here.}.
	Its stationary distribution $\pi$ is uniform, the conditional probabilities are given by
	\begin{equation}
	  \proba {ab}(t) =
	  \begin{cases}
	    \frac{1}{4} + \frac{3}{4}e^{-4t/3} =: \probadiag, & \text{if}\ a=b \\
	    \frac{1}{4} - \frac{1}{4}e^{-4t/3} =: \probaoffdiag, & \text{otherwise},
	  \end{cases}
	\end{equation}
	while the infinitesimal generator $Q$ takes the value $-1$ on the diagonal and $1/3$ elsewhere.
	\end{subsection}

	\begin{subsection}{Comparison with the likelihood objective}
	For fixed homologous sequences, the likelihood of a particular tree is
	the probability of the sequence data being generated by that tree
	(according to the chosen mutation model).
	The generative process begins at a chosen root, and proceeds along the
	unique paths leading to the leaves (which represent the taxa).
	The likelihood function is evaluated for a particular choice of tree
	topology and of branch lengths.
	The topology and branch lengths correspond respectively to the combinatorial and
	smooth aspects of the optimisation undertaken in maximum likelihood
	tree search.

	By contrast, in the approach presented here, the objective function
	depends only on smooth parameters, namely the positions of the points
	on the hyperboloid, without reference to a tree topology.
	In particular, the form of the objective does not concern paths from a
	root to the leaves and does not correspond to the generative model underlying
	tree likelihood.

	It is however naturally desirable that the objective function mimics the tree
	likelihood as closely as possible. Holder and Steel \cite{HolderSteel2011}
	show that the generating tree is always recovered as the maximiser of
	$\objective$ for sufficiently long sequence data.
	More concretely, Figure \ref{compare-objectives} shows a comparison of objective
	function $\objective$ and the likelihood objective for trees.
	An optimisation of $\objective$ is halted periodically, the pairwise
	distances between the points on the hyperboloid representing the taxa are computed, and Weighbor is used to compute
	a tree topology given these distances. Finally, the branch lengths
	of this tree are tuned to find the maximal log-likelihood for this tree
	topology.
	The three curves of Figure \ref{compare-objectives} track the development of the objective $\objective$,
	of the log likelihood of the tree obtained in the manner described, and of
	the Robinson-Foulds distance \cite{RobinsonFoulds} of that tree topology to
	that of the generating tree.
	It is apparent that maximising the objective $\objective$
	indeed has the desired effect of indirectly maximising the log-likelihood
	in tree space.

	\begin{figure}
		\centering
		\includegraphics[width=\textwidth]{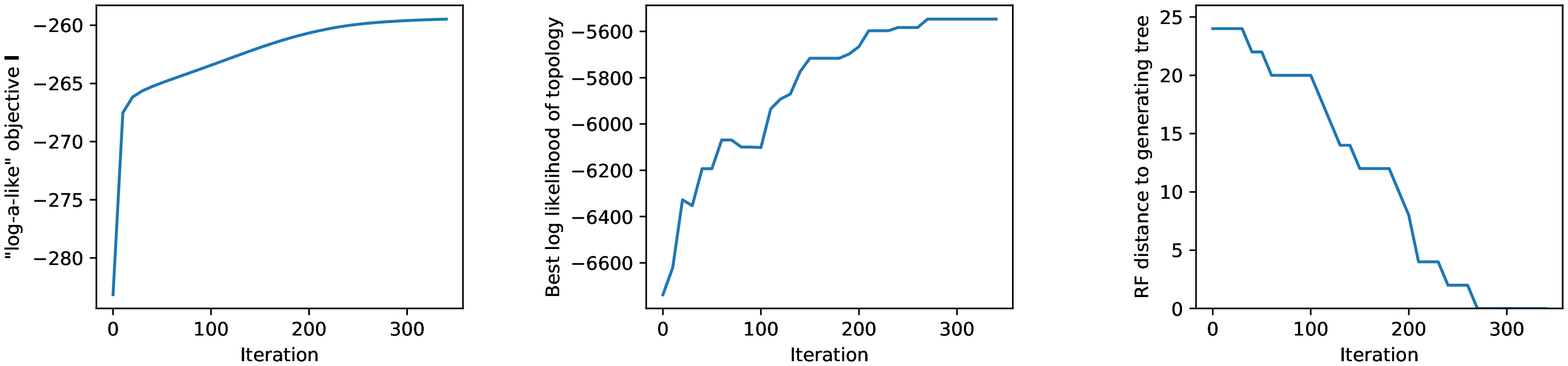}
		\caption{
		Comparison of the objective $\objective$ and the
		log-likelihood objective for trees.  Every 10 iterations, a tree
		topology is inferred using Weighbor, and the best log-likelihood of
		that topology is calculated by varying the branch lengths.
		Sequence data was generated according to the Jukes-Cantor 
		model.  The generating tree had $15$ leaves and edges sampled
		uniformly from $[0.05, 0.2]$, the sequence length was $400$,
		and our method was run with $\rho=0.5$ in dimension $10$, beginning from a
		random point configuration.
		}
		\label{compare-objectives}
	\end{figure}
	\end{subsection}

	\begin{subsection}{Optimisation}\label{Subsection:optimisation}
	Gradient ascent is used to maximise the objective
	$\objective$, following the general approach described in
	\cite{WilsonLeimeister}.
	The points of the configuration are iteratively adjusted to maximise the objective function.
	This is achieved by, for each point, considering the objective as a
	function solely of that point (i.e. considering the position of the
	other points to be fixed), computing the gradient of this function on
	the hyperboloid, and then shifting the point in the direction of the
	gradient vector.
	This ``shifting'' is achieved by applying the exponential function \eqref{Equation:exponential} (at
	the current position of the point) to a small multiple $\alpha$ of the gradient
	vector (which belongs to the tangent space at that point).
	This moves the point a distance of $\alpha$ times the norm of the gradient.
	This distance is the {\it step size} of the update.
	The multiple $\alpha$ of the gradient vector is the {\it learning
	rate}, which we take to be fixed at $\alpha=0.1$ throughout the optimisation.
	
	For the numerical stability of the optimisation a maximum step size of
	$0.05$ is used, and steps longer than this limit are truncated.
	The optimisation runs until convergence, which is deemed to have occurred
	when, in a single iteration, no point moves a distance greater than
	$5\times10^{-5}$.

	The initial placement of the points results from embedding a rooted tree.  The tree is obtained by applying Weighbor to the given sequence data, tuning the edge lengths of the resulting tree for optimal likelihood, and rooting at the midpoint.
	The embedding of the tree is reminiscent of that described in \cite{Sarkar} and generalised in \cite{Sala_2018}.
	It proceeds as follows.  Firstly, the root of the tree is placed at the basepoint of the hyperboloid.
Children of the root are placed by applying the exponential map $\Exp$ to tangent vectors (in the tangent space the point corresponding to their parent, the root) which are uniformly spaced around the unit circle of a randomly chosen 2 dimensional subspace, before being scaled by the length of their connecting edges.
Recursively, children of non-root nodes are then placed in the same manner, except that the random 2-dimensional subspace is chosen to include the tangent vector corresponding to the parent (i.e. the logarithm of the parent's point) and the uniform spacing around the unit circle is such as to accommodate this extra tangent vector.
Finally, the embeddings of all internal nodes are discarded, while the embeddings of the leaves are adopted as initial positions for the optimisation.
	\end{subsection}

	\begin{subsection}{The gradient of the objective}
	Gradient ascent is used to maximise the objective $\objective$.
	This subsection presents a formula for the gradient of $\objective$ with respect to each
	point that is required for this purpose.

	For any point $y \in \hyperboloid \rho m$, define the function
	\begin{equation}\label{distance-function-single-variable}
	\distfnsinglevar{y} : \hyperboloid \rho m \rightarrow \reals, \quad x \mapsto \dist (x, y), \quad x \in \hyperboloid \rho m.
	\end{equation}
	For any $x \in \hyperboloid \rho m$, write $\gradient x {\distfnsinglevar{y}} \in \tangentspace x {\hyperboloid \rho m}$ for the gradient at $x$.
	For any configuration $\pointconfig x = (\pointi)_{1 \leqslant i \leqslant N}$ of $N$ points on $\hyperboloid \rho m$, and for any $1 \leqslant i \leqslant N$, write
	$\objective_\pcwithouti : \hyperboloid \rho m \rightarrow \reals$ for the function $\objective$ considered as a function of the $i$th point of $\pointconfig x$ only, with all other points of the configuration held fixed:
	$$
	\objective_\pcwithouti : x \mapsto \objective ((\pcpoint{x}{1}, \dots, \pcpoint{x}{i-1}, x, \pcpoint{x}{i+1}, \dots, \pcpoint{x}{N})), \quad x \in \hyperboloid \rho m.
	$$
	For any $x \in \hyperboloid \rho m$, write $\gradient x \objective_\pcwithouti \in \tangentspace x {\hyperboloid \rho m}$ for the gradient at $x$ of this map.
	Gradient descent requires an expression for $\gradient {\pointi} \objective_\pcwithouti$.  A first step in this direction is the following:

	\begin{proposition}\label{Proposition:gradient-formula}
	Write $d_{ij} = \dist(\pointi, \pointj)$. Then:
	\begin{enumerate}
	\item\label{generalgradient} For any $\probamatrix, Q$ as in Subsection \ref{Subsection:mutation-model},
	$$ \gradient {\pointi} \objective_\pcwithouti = \tfrac{1}{L} \sum_j \sum_{\sitessigma} \frac{(Q\probamatrix (d_{ij}))_{\sigmai, \sigmaj}}{\probamatrix (d_{ij})_{\sigmai \sigmaj}} \gradient {\pointi} \distfnsinglevar {\pointj} . $$
	\item\label{jc69gradient} In the particular case of the Jukes-Cantor
	model, the gradient takes the simplified form
	$$ \gradient {\pointi} \objective_\pcwithouti =
	\sum_j e^{-4 d_{ij}/3}
	\left(
		\frac{r_{ij}}{3 \probaoffdiag(d_{ij})} - \frac{1-r_{ij}}{\probadiag(d_{ij})}
	\right) \gradient {\pointi} \distfnsinglevar {\pointj}, $$
	where $r_{ij}$ is the rate of site difference between the taxa $i, j$.
	\end{enumerate}
	\end{proposition}

	It remains, therefore, to determine an expression of the gradient of the distance function.
	\begin{proposition}\label{Proposition:distance-gradient}
	For any $x, y \in \hyperboloid \rho m$, the gradient of the distance function is given by
	$$
	\gradient x \distfnsinglevar {y} = \frac{\rho^{-2} {\minkowskiform m x y} x - y}{\sqrt{(\rho^{-1}{\minkowskiform m x y})^2 - \rho^2}} .
	$$
	\end{proposition}

	Propositions \ref{Proposition:gradient-formula} part \ref{jc69gradient} and
	\ref{Proposition:distance-gradient} together provide the needed formula for
	the gradient $\gradient {\pointi} \objective_\pcwithouti$.
	\end{subsection}

\end{section}

\begin{section}{Implementation \& scalability}\label{Section:implementation}
	The proposed method was implemented using Python 3.7.4 and Cython 0.29.22.
	The code of both the implementation and the evaluation framework are available\footnote{\href{https://github.com/lateral/phylogeny}{https://github.com/lateral/phylogeny}}.
	The implementation, which supports only the Jukes-Cantor model, updates each
	point of the configuration in turn using the formulae of Proposition
	\ref{Proposition:gradient-formula} part \ref{jc69gradient} and Proposition
	\ref{Proposition:distance-gradient}.
	Inspection of these formulae is sufficient to determine that the time cost of
	calculating a single gradient is $O(N m)$, which dominates the $O(m)$  cost
	of then updating that point, given the gradient, using formula
	\eqref{Equation:exponential}.
	Thus the total time cost of a single iteration is $O(N^2 m)$.
	Iterations continue until no point has travelled a distance greater than $5 \times 10^{-5}$.
	This typically occurs after some thousands of iterations.
	For example, in the simulation study detailed in Subsection
	\ref{Experiment:sequence-length}, the modal number of iterations until
	convergence was $1598$, while the maximum was $2153$.
	Figure \ref{Figure:scalability} compares the time cost of a fixed number of
	iterations of the present method to the other methods considered in the
	evaluation (cf. Section \ref{Section:experiments}).
	While the present implementation is only single-threaded, we note that a
	straight-forward parallelisation is possible, where a single thread is
	responsible of the calculation of the gradient and the updating of a single
	point.

	\begin{figure}
		\centering
		\includegraphics[width=0.75\textwidth]{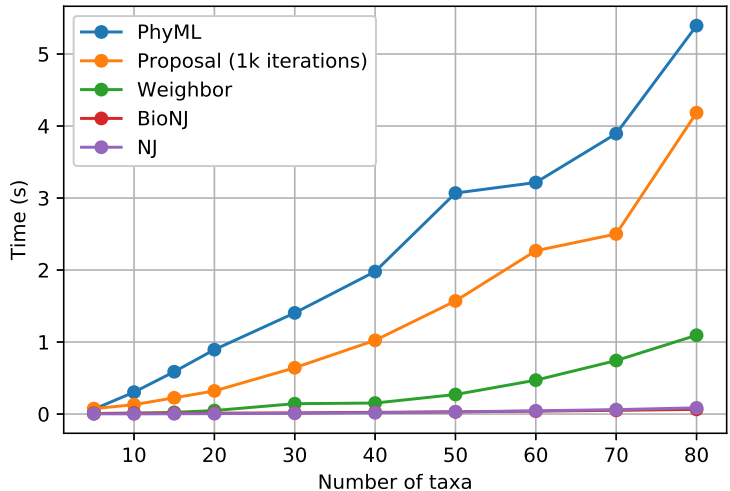}
		\caption{Time taken for 1000 iterations of the implementation (with $m=10$
		and for various numbers $N$ of taxa) compared to the baseline methods of
		the evaluation (cf. Section \ref{Section:experiments}).}
		\label{Figure:scalability}
	\end{figure}

\end{section}

\begin{section}{Simulation studies}\label{Section:experiments}
	This section presents two simulation studies comparing the proposed method to
	other tree inference methods, under varying conditions.
	Three distance-based methods were chosen for comparison:
	Weighbor\footnote{Version 1.2, obtained from
	\href{http://www.t6.lanl.gov/billb/weighbor/download.html}
	{http://www.t6.lanl.gov/billb/weighbor/download.html}
	via the WayBackMachine.} \cite{Weighbor}, BIONJ\footnote{Obtained from
	\href{http://www.atgc-montpellier.fr/bionj/download.php}
	{http://www.atgc-montpellier.fr/bionj/download.php}.}
	\cite{bionj} and neighbor-joining (NJ)\footnote{Implemented in
	\href{https://dendropy.org/}{DendroPy\cite{DendroPy}, version 4.4.0}.} \cite{NJ}.
	PhyML\cite{PhyML}\footnote{Version 3.1,
	obtained
	\href{http://www.atgc-montpellier.fr/phyml/versions.php}
	{http://www.atgc-montpellier.fr/phyml/versions.php}.}
	is also included as a representative of the various maximum likelihood tree search methods.
	Subsection \ref{Experiment:number-of-taxa} considers the effect of varying
	the number of taxa, while subsection \ref{Experiment:sequence-length} reports
	on performance as the length of the homologous sequence data is varied.

	\begin{subsection}{Inference of tree topologies}
	The performance of each method is assessed for a given tree and homologous
	sequence data generated from that tree under the Jukes-Cantor model.  The
	distance-based methods take as input the estimated
	pairwise distances, along with the length of the sequences (in the case of
	Weighbor), and produce a tree topology.  These pairwise distances are
	estimated in the usual fashion (that is, independently from one another,
	using the maximum-likelihood criterion) from the observed rates of site
	difference.  The proposed method takes as input the rates of site
	difference, and is run until convergence. The pairwise distances are
	then measured from the learned point configuration.  Weighbor is then used to
	obtain a tree topology from these pairwise distances (this is distinct from
	the use of Weighbor as a compared method).
	PhyML takes the sequence data as input and is set to use nearest-neighbour
	interchange to move among the possible tree topologies.
	\end{subsection}

	\begin{subsection}{Evaluation metrics}
	The tree topologies resulting from each method are evaluated according
	to two metrics.  The first is the rate of matching, or exceeding, the
	likelihood of the generating tree.  For each resulting tree topology,
	PAUP*\footnote{Version {\tt 4.0a165}, obtained from
	\href{http://phylosolutions.com/}{http://phylosolutions.com/}.} \cite{paup}
	is used to find the maximal likelihood of a tree with that tree topology,
	given the sequence data, by tuning the branch lengths.  This is then compared
	to the likelihood of the generating tree.  The method under evaluation is deemed to have
	succeeded in its inference if the likelihood of the inferred tree is at least as
	high as the likelihood of the generating tree, and the rate at which this
	occurs over a collection of trees and sequences is reported as an evaluation metric.  This
	metric is particularly useful in cases where the sequence length is very
	short, meaning that the generating tree often does not have the highest
	likelihood.

	The second metric is the average rate of coincidence of the inferred tree
	topology and the generating tree topology.
	\end{subsection}

	\begin{subsection}{Generation of trees and sequences}
	For a given number of leaves, an unrooted tree topology is drawn uniformly at random using a pure birth process.
	The tree topology becomes an unrooted tree by sampling edge lengths uniformly
	at random from the interval $[0.05, 0.2]$, and this tree is then used to
	generate sequences of the specified length using the Jukes-Cantor model.
	\end{subsection}

	\begin{subsection}{Choice of hyperparameters}
	In the simulation studies reported here, the proposed method is run until
	convergence at learning rate $0.1$, with a maximum step size of $0.05$, on
	points initialised as described in subsection \ref{Subsection:optimisation}.
	The hyperboloid of radius $\rho=0.5$ and dimension $30$ was used.
	\end{subsection}

	\begin{subsection}{Simulation study: variation of the number of taxa}\label{Experiment:number-of-taxa}
	This simulation study compares the performance of the methods at the inference of
	trees as the number of taxa varies between $5$ and $100$.
	Tree inference is more difficult with a higher number of taxa, firstly since
	there are more trees to choose from, and secondly since it results (under the
	chosen tree generation regime) in longer inter-leaf distances, which are more
	difficult to estimate from finite-length sequence data. 
	For each number of taxa under consideration, $10$ trees were drawn at random, and for each tree
	$4$ sets of homologous sequence data of length $200$ were generated.

	The results are depicted in Figure \ref{Figure:number-of-leaves}.
	The performance of all methods is optimal when the number of taxa is very
	small ($5$), and deteriorates as the number of taxa increases.
	PhyML performs almost perfectly throughout.
	Weighbor performs best among the distance-based methods.  The proposed method performs
	comparably with Weighbor when the number of taxa is small, and significantly
	better than all distance-based methods as the number of taxa grows.

		\begin{figure}[htb]%
			\begin{minipage}[t]{0.48\textwidth}
			\centering
			\includegraphics[width=\textwidth]{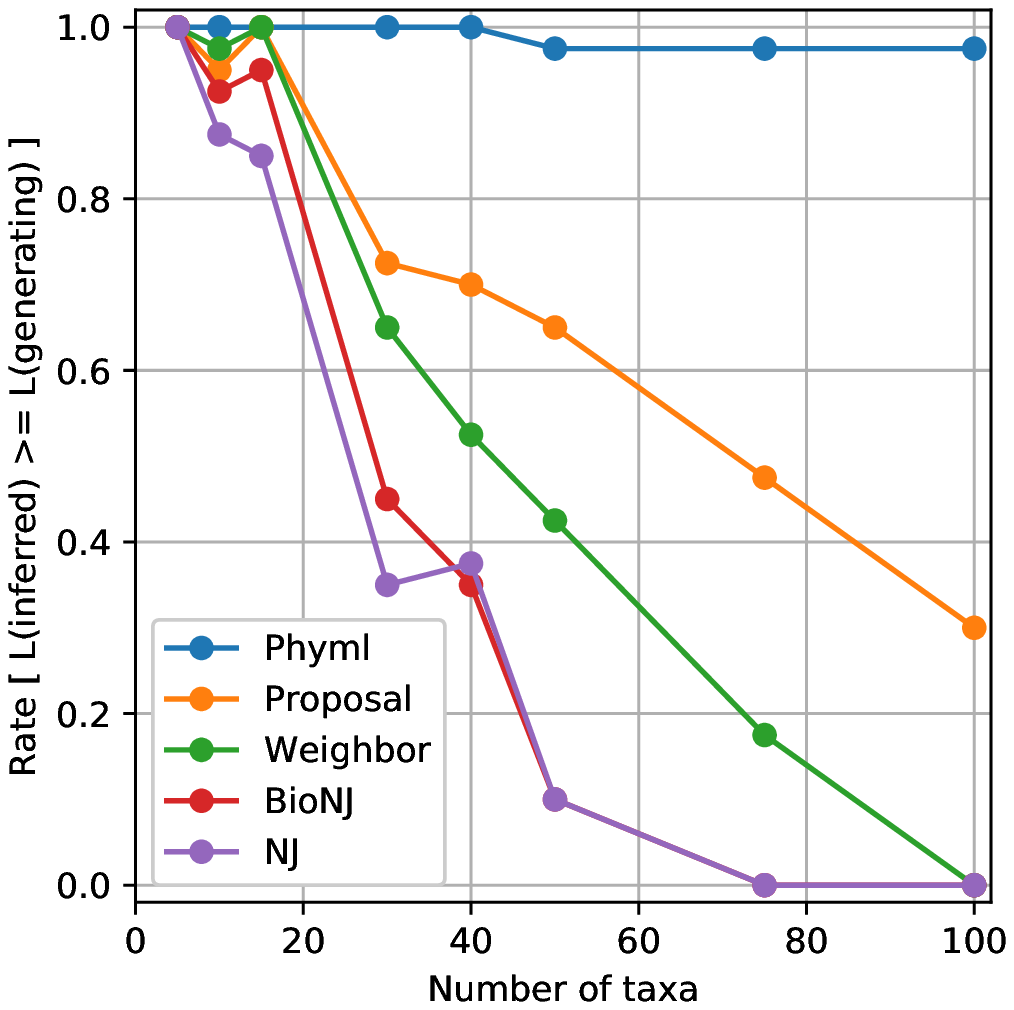}
			\subcaption{}\label{number-of-leaves-likelihood-wrt-generating-article}
			\end{minipage}%
			\hfill
			\begin{minipage}[t]{0.48\textwidth}
			\centering
			\includegraphics[width=\textwidth]{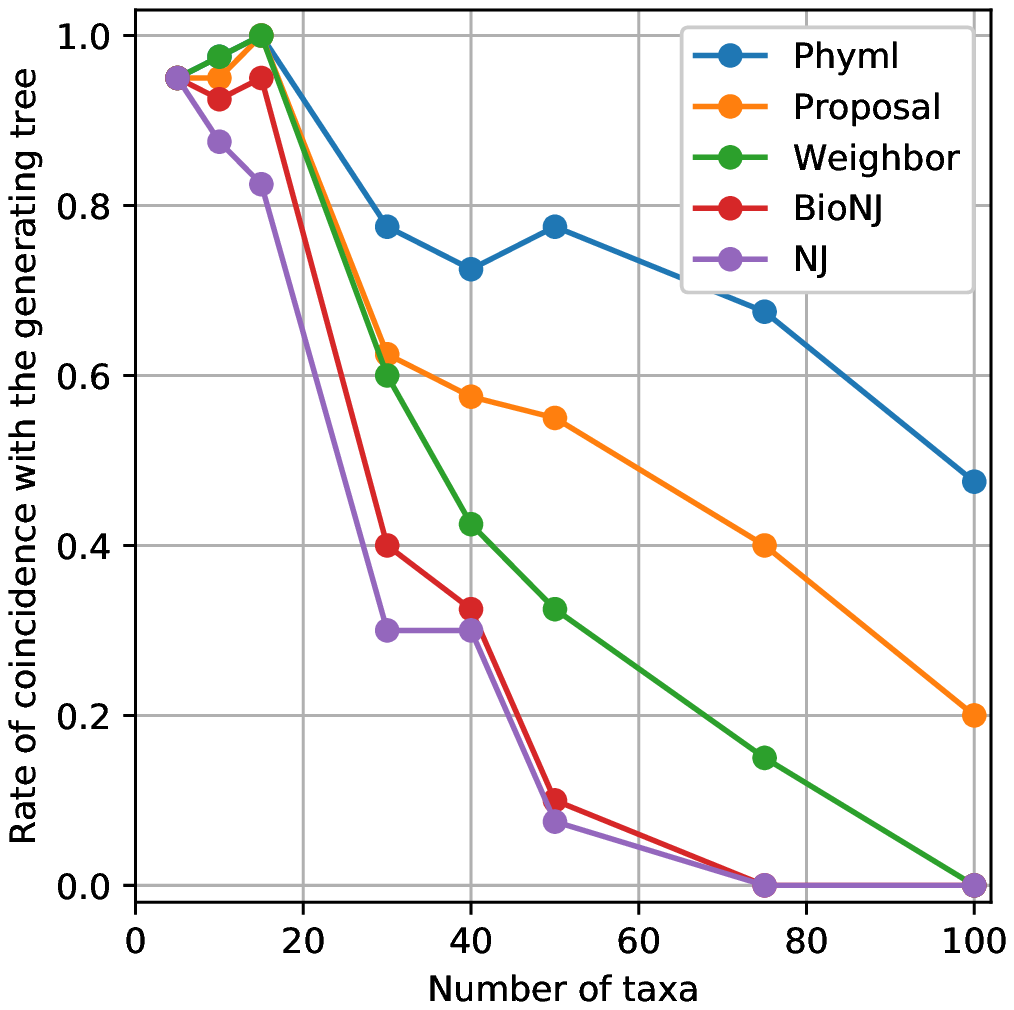}
			\subcaption{}\label{number-of-leaves-topological-accuracy-article}
			\end{minipage}%
			\caption{
			Performance of the compared methods as the number of taxa is varied.
			The proposed approach performs comparably when the number
			of taxa is small, and significantly better than the distance-based
			methods when the number of taxa is large.
			Figure \ref{number-of-leaves-likelihood-wrt-generating-article} shows the
			rate at which the likelihood of the inferred tree is at least as high as
			the likelihood of the generating tree, while Figure
			\ref{number-of-leaves-topological-accuracy-article} plots the rate of
			coincidence of inferred- and generating- tree topologies.
			}
			\label{Figure:number-of-leaves}
		\end{figure}

	\end{subsection}

	\begin{subsection}{Simulation study: variation of sequence length}\label{Experiment:sequence-length}
	This simulation study compares the performance of the methods as the length of the
	homologous sequence data is varied over a range of values from $100$ to $750$
	bases.
	As the generating tree can be recovered reliably by algorithms such as
	neighbor-joining when the sequence data is sufficiently large, the shorter
	lengths are of particular interest in the comparison.
	Homologous sequences were generated $12$ times from $8$ randomly drawn trees
	with $30$ leaves.

	The results are depicted in Figure \ref{Figure:sequence-length}.
	As expected, all methods perform better as the sequence length increases and sampling noise consequently decreases.
	PhyML performs almost perfectly, even for very short sequences.
	In all cases, and particularly in the case of very short sequences,
	the proposed method performs favourably in comparison to the distance-based methods.
	This contrast in performance with the distance based methods in the case of very short sequences is expected,
	as the proposed approach effects a joint estimation of the pairwise
	distances.  The proposed method thus has more signal available in the determination of
	each pairwise distance than the distance methods, which take as input the
	matrix of independently estimated pairwise distances.

		\begin{figure}[htb]%
			\begin{minipage}[t]{0.48\textwidth}
			\centering
			\includegraphics[width=\textwidth]{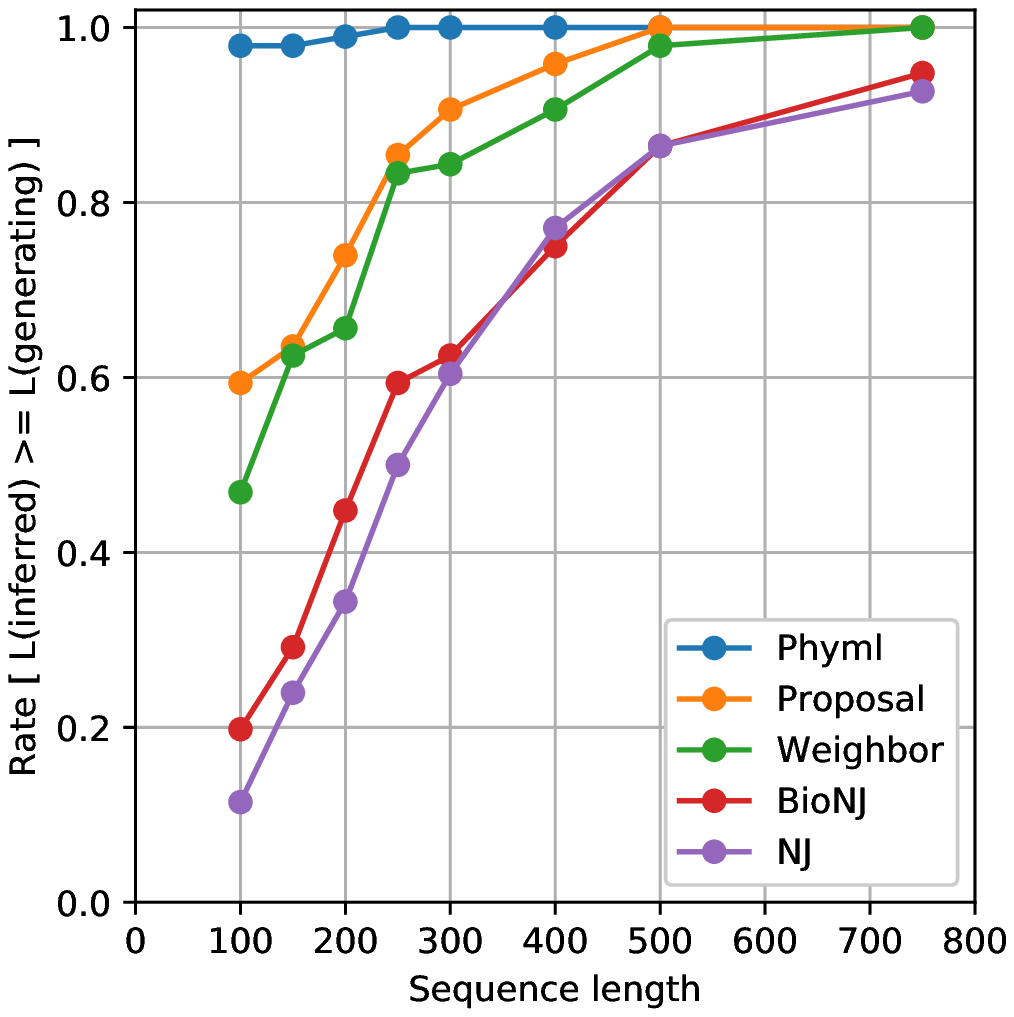}
			\subcaption{}\label{sequence-length-experiment-ll-article}
			\end{minipage}%
			\hfill
			\begin{minipage}[t]{0.48\textwidth}
			\centering
			\includegraphics[width=\textwidth]{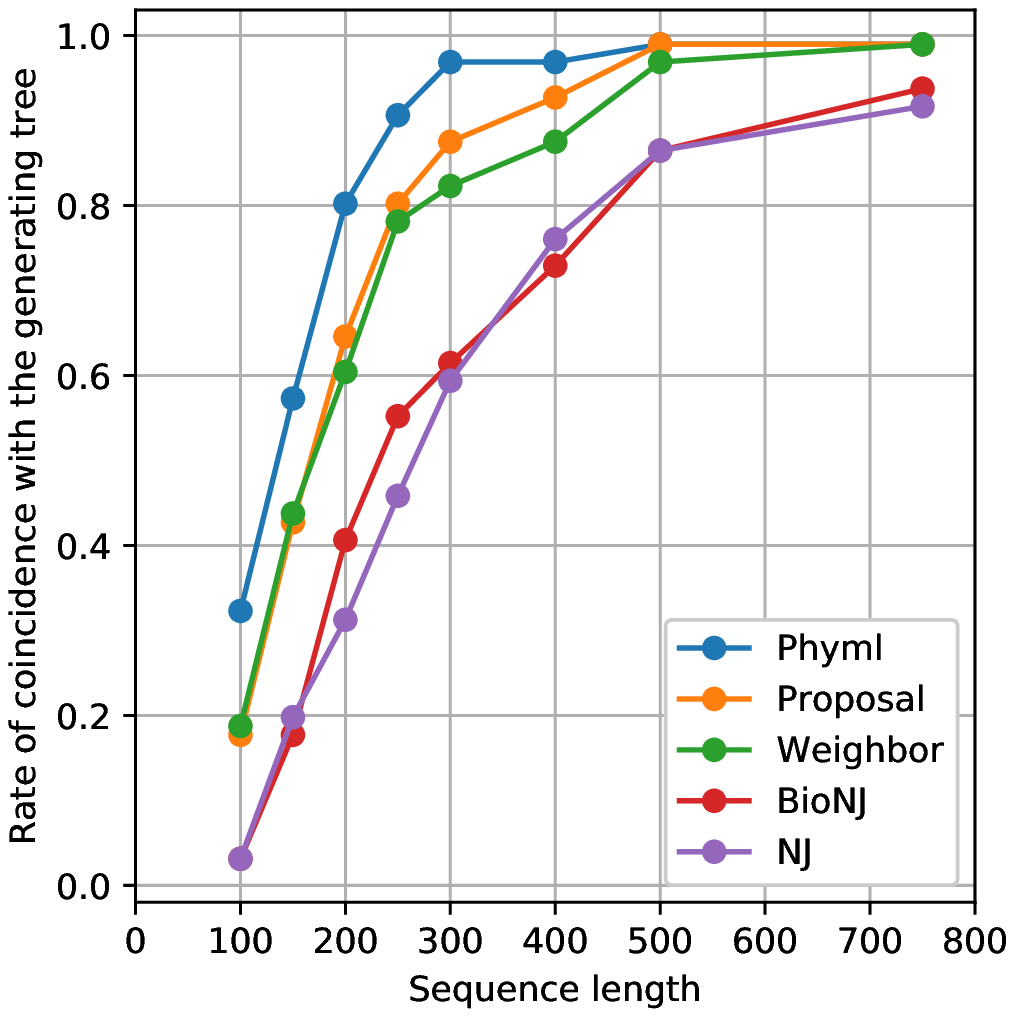}
			\subcaption{}\label{sequence-length-experiment-topo-coincidence-article}
			\end{minipage}%
			\caption{
			Performance of the compared methods as the sequence length is varied.
			The proposed approach performs favourably in all cases,
			and particularly in the case of short sequences.
			Figure \ref{sequence-length-experiment-ll-article} shows the
			rate at which the likelihood of the inferred tree is at least as high as
			the likelihood of the generating tree, while Figure
			\ref{sequence-length-experiment-topo-coincidence-article} plots the rate of
			coincidence of inferred- and generating- tree topologies.
			}
			\label{Figure:sequence-length}
		\end{figure}

	\end{subsection}

\end{section}

\begin{section}{Discussion and open questions}\label{Section:discussion}
This paper contributes the idea that tree metrics can be estimated
using point configurations in hyperbolic space and an objective function designed to mimic the
likelihood on tree space.
The paper suggests several possible directions for future research:

\begin{enumerate}
\item A goal of great theoretical interest would be the definition of a distance between two unrooted trees.
This might be achieved as the limit as $\rho \rightarrow 0$ of the distance between point configurations on $\hyperboloid \rho m$ that maximise the objective $\objective$.  The computation of such distances between point configurations would need to take into account the isometries discussed in Section \ref{Section:general-approach}.
\item It would be of interest to determine if the method proposed here is biased in the topologies of the trees that it infers (a question raised by Allen Rodrigo).
\item From a practical standpoint, finite precision arithmetic limits the values of $\rho$ at which the proposed optimisation can be carried out.  Can the tree metric obtained in the theoretical limit as $\rho \rightarrow 0$ be determined via extrapolation from the metrics obtained at those values of $\rho$ that facilitate optimisation? (A suggestion of Ben Kaehler).
\item The method presented here would be improved by the development of an algorithm deriving trees directly from their corresponding hyperbolic point configurations, rather than applying Weighbor to the inter-point distances.
\item Holder and Steel \cite{HolderSteel2011} note that the pairwise likelihood is at a disadvantage to standard likelihood, in that the latter explicitly incorporates ancestral characters.  Can this deficiency be overcome, now that the pairwise likelihood is endowed with a ambient space?
\end{enumerate}

\end{section}

\bibliographystyle{alpha}
\bibliography{hyperbolic-citations}

\newcommand{\etalchar}[1]{$^{#1}$}
\begin{thebibliography}{N{\v{S}pakula}16}

\bibitem[ABC{\etalchar{+}}91]{Alonsoetal}
J.~M. Alonso, T.~Brady, D.~Cooper, V.~Ferlini, Lustig, M.~Mihalik, M.~Shapiro,
  and H.~Short.
\newblock Notes on word hyperbolic groups.
\newblock In {\em Group theory from a geometrical viewpoint}, 1991.

\bibitem[BHV01]{BHV2001}
Louis~J. Billera, Susan~P. Holmes, and Karen Vogtmann.
\newblock Geometry of the space of phylogenetic trees.
\newblock {\em Advances in Applied Mathematics}, 27(4):733 -- 767, 2001.

\bibitem[Bou20]{boumal2020intromanifolds}
Nicolas Boumal.
\newblock An introduction to optimization on smooth manifolds.
\newblock Available online, Nov 2020.

\bibitem[BS00]{HypertreeBingham2000}
Jonathan Bingham and Sucha Sudarsanam.
\newblock Visualizing large hierarchical clusters in hyperbolic space.
\newblock {\em Bioinformatics (Oxford, England)}, 16:660--1, 08 2000.

\bibitem[BSH00]{Weighbor}
William Bruno, Nicholas Socci, and Aaron Halpern.
\newblock Weighted neighbor joining: A likelihood-based approach to
  distance-based phylogeny reconstruction.
\newblock {\em Molecular biology and evolution}, 17:189--97, 02 2000.

\bibitem[Bun71]{Buneman1971}
Peter Buneman.
\newblock The recovery of trees from measures of dissimilarity.
\newblock In {\em Mathematics the the Archeological and Historical Sciences},
  pages 387--395, United Kingdom, 1971. Edinburgh University Press.

\bibitem[CC16]{CvetkovskiCrovella2016}
Andrej Cvetkovski and Mark Crovella.
\newblock Multidimensional scaling in the poincaré disk.
\newblock {\em Applied Mathematics \& Information Sciences}, 10(1):125--133,
  2016.

\bibitem[CFKP97]{Cannon}
J.~W. Cannon, W.~J. Floyd, R.~Kenyon, and W.~R. Parry.
\newblock Hyperbolic geometry.
\newblock In {\em Flavors of Geometry}, pages 59--115. Cambridge University
  Press, 1997.

\bibitem[CGCR20]{chami2020trees}
Ines Chami, Albert Gu, Vaggos Chatziafratis, and Christopher Ré.
\newblock From trees to continuous embeddings and back: Hyperbolic hierarchical
  clustering, 2020.

\bibitem[DSN{\etalchar{+}}18]{dhingra-etal-2018-embedding}
Bhuwan Dhingra, Christopher Shallue, Mohammad Norouzi, Andrew Dai, and George
  Dahl.
\newblock Embedding text in hyperbolic spaces.
\newblock In {\em Proceedings of the Twelfth Workshop on Graph-Based Methods
  for Natural Language Processing ({T}ext{G}raphs-12)}, pages 59--69, New
  Orleans, Louisiana, USA, June 2018. Association for Computational
  Linguistics.

\bibitem[Efi63]{Efimov}
N.~V. Efimov.
\newblock Impossibility of a complete regular surface in euclidean 3-space
  whose gaussian curvature has a negative upper bound.
\newblock {\em Sov. Math. (Doklady)}, 4:843–846, 1963.

\bibitem[Gas97]{bionj}
O.~Gascuel.
\newblock {{B}{I}{O}{N}{J}: an improved version of the {N}{J} algorithm based
  on a simple model of sequence data}.
\newblock {\em Mol. Biol. Evol.}, 14(7):685--695, Jul 1997.

\bibitem[Gas07]{MathEvoPhyl}
Olivier Gascuel.
\newblock {\em Mathematics of Evolution and Phylogeny}.
\newblock Oxford University Press, Inc., New York, NY, USA, 2007.

\bibitem[GBH18]{Ganea2018}
Octavian Ganea, Gary Becigneul, and Thomas Hofmann.
\newblock Hyperbolic neural networks.
\newblock In S.~Bengio, H.~Wallach, H.~Larochelle, K.~Grauman, N.~Cesa-Bianchi,
  and R.~Garnett, editors, {\em Advances in Neural Information Processing
  Systems}, volume~31. Curran Associates, Inc., 2018.

\bibitem[GDL{\etalchar{+}}10]{PhyML}
Stéphane Guindon, Jean-François Dufayard, Vincent Lefort, Maria Anisimova,
  Wim Hordijk, and Olivier Gascuel.
\newblock {New Algorithms and Methods to Estimate Maximum-Likelihood
  Phylogenies: Assessing the Performance of PhyML 3.0}.
\newblock {\em Systematic Biology}, 59(3):307--321, 05 2010.

\bibitem[HHL04]{WalrusHughes2004}
Timothy Hughes, Young Hyun, and David Liberles.
\newblock Visualising very large phylogenetic trees in three dimensional
  hyperbolic space.
\newblock {\em BMC bioinformatics}, 5:48, 05 2004.

\bibitem[HS11]{HolderSteel2011}
Mark~T. Holder and Mike Steel.
\newblock Estimating phylogenetic trees from pairwise likelihoods and posterior
  probabilities of substitution counts.
\newblock {\em Journal of Theoretical Biology}, 280(1):159--166, 2011.

\bibitem[Joh14]{Johnson2014}
Christopher~C. Johnson.
\newblock Logistic matrix factorization for implicit feedback data.
\newblock {\em Advances in Neural Information Processing Systems}, 27(78):1--9,
  2014.

\bibitem[LC78]{LindmanCaelli}
Harold Lindman and Terry Caelli.
\newblock Constant curvature riemannian scaling.
\newblock {\em Journal of Mathematical Psychology}, 17(2):89 -- 109, 1978.

\bibitem[Lou20]{Loustau2020hyperbolic}
Brice Loustau.
\newblock Hyperbolic geometry, 2020.

\bibitem[LW18]{LeimeisterWilson}
Matthias Leimeister and Benjamin~J. Wilson.
\newblock Skip-gram word embeddings in hyperbolic space.
\newblock {\em CoRR}, abs/1809.01498, 2018.

\bibitem[MMF21]{Matsumoto2021}
Hirotaka Matsumoto, Takahiro Mimori, and Tsukasa Fukunaga.
\newblock {Novel metric for hyperbolic phylogenetic tree embeddings}.
\newblock {\em Biology Methods and Protocols}, 6(1), 03 2021.
\newblock bpab006.

\bibitem[{Mun}97]{Munzner1998}
T.~{Munzner}.
\newblock H3: laying out large directed graphs in 3d hyperbolic space.
\newblock In {\em Proceedings of VIZ '97: Visualization Conference, Information
  Visualization Symposium and Parallel Rendering Symposium}, pages 2--10, 1997.

\bibitem[MZS{\etalchar{+}}19]{Monath2019}
Nick Monath, Manzil Zaheer, Daniel Silva, Andrew McCallum, and Amr Ahmed.
\newblock Gradient-based hierarchical clustering using continuous
  representations of trees in hyperbolic space.
\newblock In {\em The 25th ACM SIGKDD Conference on Knowledge Discovery and
  Data Mining (KDD ’19)}, 2019.

\bibitem[NK18]{NickelKiela2018}
Maximilian Nickel and Douwe Kiela.
\newblock Learning continuous hierarchies in the lorentz model of hyperbolic
  geometry.
\newblock In {\em Proceedings of the 35th International Conference on Machine
  Learning, {ICML} 2018, Stockholmsm{\"{a}}ssan, Stockholm, Sweden, July 10-15,
  2018}, pages 3776--3785, 2018.

\bibitem[N{\v{S}pakula}16]{BogdanHyperbolicity}
Bogdan {Nica} and J\'an {\v{S}pakula}.
\newblock {Strong hyperbolicity}.
\newblock {\em {Groups Geom. Dyn.}}, 10(3):951--964, 2016.

\bibitem[Pol20]{Poleksic2020}
Aleksandar Poleksic.
\newblock Hyperbolic matrix factorization reaffirms the negative curvature of
  the native biological space.
\newblock {\em bioRxiv}, 2020.

\bibitem[RF81]{RobinsonFoulds}
D.F. Robinson and L.R. Foulds.
\newblock Comparison of phylogenetic trees.
\newblock {\em Mathematical Biosciences}, 53(1):131--147, 1981.

\bibitem[Sar12]{Sarkar}
Rik Sarkar.
\newblock Low distortion delaunay embedding of trees in hyperbolic plane.
\newblock In Marc van Kreveld and Bettina Speckmann, editors, {\em Graph
  Drawing}, pages 355--366, Berlin, Heidelberg, 2012. Springer Berlin
  Heidelberg.

\bibitem[SDSGR18]{Sala_2018}
Frederic Sala, Chris De~Sa, Albert Gu, and Christopher Re.
\newblock Representation tradeoffs for hyperbolic embeddings.
\newblock In {\em Proceedings of the 35th International Conference on Machine
  Learning}, pages 4460--4469, Stockholmsmässan, Stockholm Sweden, 10--15 Jul
  2018.

\bibitem[SH10]{DendroPy}
Jeet Sukumaran and Mark~T. Holder.
\newblock {DendroPy: a Python library for phylogenetic computing}.
\newblock {\em Bioinformatics}, 26(12):1569--1571, 04 2010.

\bibitem[SN87]{NJ}
N.~Saitou and M.~Nei.
\newblock The neighbor-joining method: a new method for reconstructing
  phylogenetic trees.
\newblock {\em Molecular biology and evolution}, 4 4:406--25, 1987.

\bibitem[Swo01]{paup}
David~L. Swofford.
\newblock Paup*: Phylogenetic analysis using parsimony (and other methods)
  version 4.0 beta, 2001.

\bibitem[Tav86]{TavareGTR}
S.~Tavar{\'e}.
\newblock Some probabilistic and statistical problems in the analysis of dna
  sequences.
\newblock 1986.

\bibitem[TBG19]{DBLP:conf/iclr/TifreaBG19}
Alexandru Tifrea, Gary B{\'{e}}cigneul, and Octavian{-}Eugen Ganea.
\newblock Poincare glove: Hyperbolic word embeddings.
\newblock In {\em 7th International Conference on Learning Representations,
  {ICLR} 2019, New Orleans, LA, USA, May 6-9, 2019}. OpenReview.net, 2019.

\bibitem[WHPD14]{WilsonHancock}
R.~C. Wilson, E.~R. Hancock, E.~Pekalska, and R.~P.~W. Duin.
\newblock Spherical and hyperbolic embeddings of data.
\newblock {\em IEEE Transactions on Pattern Analysis and Machine Intelligence},
  36(11):2255--2269, Nov 2014.

\bibitem[WL18]{WilsonLeimeister}
Benjamin Wilson and Matthias Leimeister.
\newblock {Gradient descent in hyperbolic space}.
\newblock {\em arXiv:1805.08207}, 2018.

\bibitem[Zar65]{Zaretskii1965}
K.~A. Zaretskii.
\newblock Constructing a tree on the basis of a set of distances between the
  hanging vertices.
\newblock {\em Uspekhi Mat. Nauk}, 20(6(126)):90--92, 1965.

\end{thebibliography}

\newpage
\section*{Appendix: Related work}\label{Section:related-work}
Hyperbolic space is well-suited for the representation of trees, owing to
the surface area and volume of any sphere being exponential in the sphere radius.  Indeed,
hyperbolic space has been employed in the software package H3
\cite{Munzner1998} for the visualisation of trees, and by packages Hypertree
\cite{HypertreeBingham2000} and Walrus \cite{WalrusHughes2004} for the
visualisation of phylogenetic trees, in particular.  An algorithm for the
construction of trees in hyperbolic plane is given in \cite{Sarkar}, while the
higher dimensional case is tackled in \cite{Sala_2018}.  A machine
learning approach for the embedding of trees is presented in
\cite{NickelKiela2018}, in the context of linguistics and social networks.
The work \cite{Matsumoto2021} investigates the embedding of trees in a
phylogenetic context using a transformation of the inter-node distance.
This latter work concerns the learning of an embedding of a prescribed
tree, whereas the present work concerns the learning of the tree itself, via
an embedding, from given sequence data.

This work considers the problem of learning point configurations in hyperbolic
space to reflect the mutation distance between taxa.  In the linguistic field
of distributional semantics, the learning of point configurations has been
explored in \cite{LeimeisterWilson, dhingra-etal-2018-embedding,
DBLP:conf/iclr/TifreaBG19}.  The related problem of multidimensional scaling in
hyperbolic space has also been treated in \cite{LindmanCaelli,WilsonHancock,
Sala_2018, CvetkovskiCrovella2016}.  The techniques required for many further
applications of machine learning in hyperbolic space are developed in
\cite{Ganea2018}.

In the present work, a tree is derived from a point configuration by applying a
distance-based method.  It is to be hoped, however, that future work will
construct the tree in hyperbolic space using the point configuration directly.
Pertinent to this challenge will be the hyperbolic hierarchical clustering
algorithms presented in \cite{chami2020trees,Monath2019}.

The work \cite{Poleksic2020} seeks to uncover a latent biological space using
an adaptation of logistic matrix factorisation \cite{Johnson2014} to hyperbolic
space.  Specifically, the objective function of logistic matrix factorisation
is modified by replacing the dot product (which approximates distance on the
sphere, in Euclidean space) by the bilinear form on Minkowski space.  However,
given the highly non-linear relationship between the latter and distance in
hyperbolic space (see equation \eqref{hyperboloiddistance}), the connection of their method to
hyperbolic geometry is unclear.

A point configuration can be viewed as a single point on a product manifold
formed from multiple copies of the original manifold.  Thus the approach
presented here represents the trees with a given number of leaves as points in
a space (the product manifold),
and in this sense is related to the well-studied space of Billera, Holmes, and
Vogtmann (BHV space)\cite{BHV2001}.  Our approach differs from theirs in
several ways.  BHV space is a polytope, but not a manifold, and does not
possess the structure required to perform numeric optimisation.  In particular,
the construction of BHV space explicitly encodes the combinatorial optimisation
landscape of maximum likelihood tree search in the joining of the quadrants
representing each tree.  By contrast, the nearness of trees in our approach
arises naturally from the geometry of hyperbolic space.  In addition, in BHV
space, each tree is represented by a single point on the polytope.  However, in
our approach, a multitude of points in the product manifold represent the same
tree, resulting from isometries of the space underlying the product manifold.
Finally, points in BHV space represent tree metrics, whereas points on our
product manifold are only approximately tree metrics (cf. Section
\ref{Section:four-point-condition}).

\section*{Appendix: Hyperbolic space}\label{Section:hyperbolic-space}
Hyperbolic $m$-space is an $m$-dimensional Riemannian manifold.
Like Euclidean space, it is unbounded in its extent and homogeneous (``isotropic'').
Unlike Euclidean space, the circumference of a circle grows exponentially in the radius.
This property makes hyperbolic space particularly well-suited for the embedding of trees, since the rate of growth of the circumference is able to match the rate of growth of the number of nodes as the depth of the tree increases.

Hyperbolic space can not be embedded without distortion in Euclidean space \cite{Efimov}, but there are various {\it models} of hyperbolic space that allow calculations to be carried out.
This paper employs the Poincaré model for visualisation and the the hyperboloid model for optimisation.
This section provides a brief review of these two models.  The reader is referred to \cite{Cannon} for further background in hyperbolic geometry and to \cite{WilsonLeimeister} for a guide to optimisation using the hyperboloid model.

\subsection*{The Poincaré ball}
\begin{figure}[htb]%
	\begin{minipage}[t]{0.45\textwidth}
	\centering
	\includegraphics[width=\textwidth]{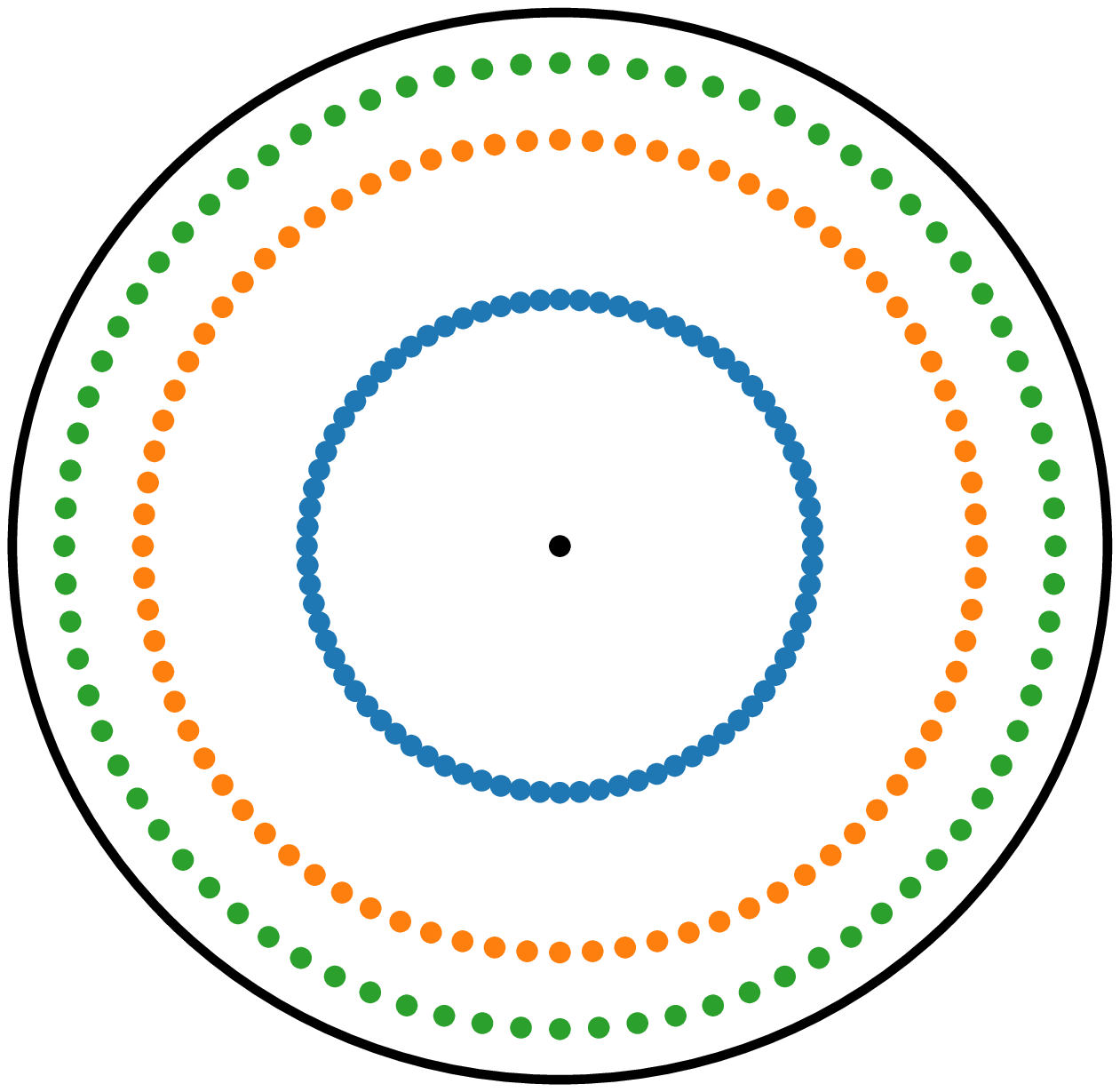}
	\subcaption{}\label{poincare-disc-circles-1}
	\end{minipage}%
	\hfill
	\begin{minipage}[t]{0.45\textwidth}
	\centering
	\includegraphics[width=\textwidth]{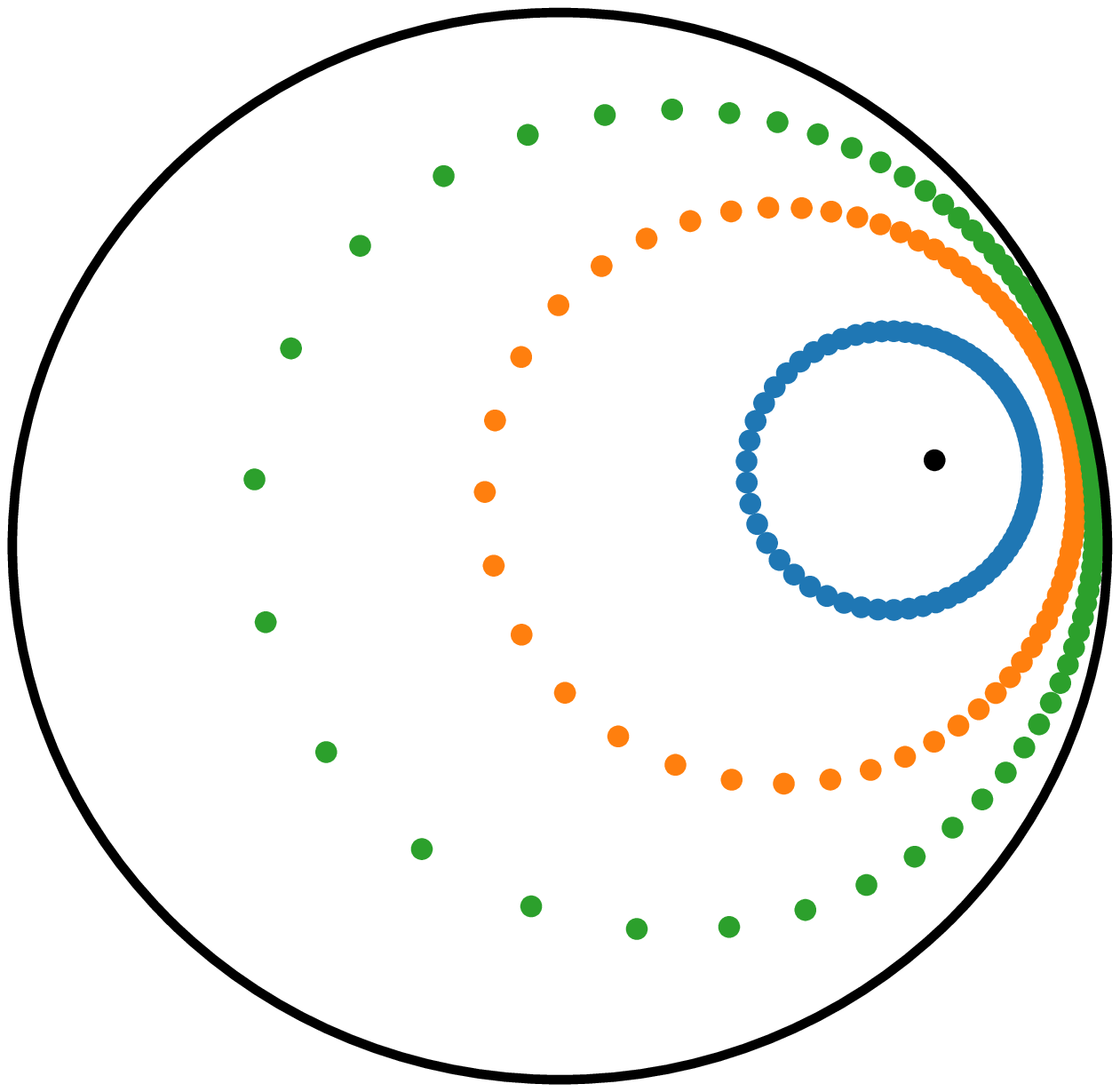}
	\subcaption{}\label{poincare-disc-circles-2}
	\end{minipage}%
	\caption{Illustrations of distance function \eqref{pbdistance} on the Poincaré disc $\pb 1 2$. Both figures depict three concentric circles of with radii $1, 2, 3$, each represented by $80$ points placed at regular intervals around the perimeter.  In Figure \ref{poincare-disc-circles-1}, the centre of the circles coincides with the centre of the Poincaré disc.  In Figure \ref{poincare-disc-circles-2}, this is not the case.}  
	\label{poincare-disc-circles}
\end{figure}

The Poincaré ball is a model of hyperbolic space convenient for visualisation.
The Poincaré ball model represents $m$-dimensional hyperbolic space as the interior of the ball of radius $\rho > 0$ in $m$-dimensional Euclidean space:
$$ \pb \rho m = \{\, \asvector x \in \reals^m \mid \| \asvector x \| < \rho \,\}.$$
(The meaning of the parameter $\rho$ will be discussed later.)
In this model, the distance between two points can be calculated using the ambient Euclidean geometry via
\begin{equation}\label{pbdistance}
	\dist_{\pb \rho m} (\asvector x, \asvector y) = \rho \arccosh \left(1 + 2 \frac{ \rho^2 \| \asvector x - \asvector y \|^2}{(\rho^2 - \| \asvector x \|^2)(\rho^2 - \| \asvector y \|^2)}\right), \quad \asvector x, \asvector y \in \pb \rho m.
\end{equation}
In the case where $m=2$, the Poincaré ball (disc) model permits a very convenient (if distorted) visualisation of hyperbolic space.
This distortion manifests itself in the dilation of distance between a pair of points by the closeness of either to the Euclidean boundary of the ball, and consequently in the seemingly distinguished nature of the centre point (whereas the space is in fact isotropic, meaning there are no distinguished points).
The peculiarities of the distance function on the Poincaré disc are illustrated in Figure \ref{poincare-disc-circles}.

\subsection*{The hyperboloid model}
Hyperbolic space can be constructed, analogously to the sphere in its Euclidean ambient, as a pseudo-sphere (or hyperboloid) in a linear ambient space called {\it Minkowski space}.
This construction is called the ``hyperboloid model'' of hyperbolic space, and is very convenient for numerical optimisation \cite{WilsonLeimeister}.

Write $\minkowski m$ for a copy of $\reals^{m+1}$ equipped with the bilinear form $\minkowskiform m \cdot \cdot$ given by:
$$ \textstyle \minkowskiform m {\asvector x} {\asvector y} = \left( \sum_{i=1}^m x_i y_i \right) - x_{m+1} y_{m+1}, \quad x, y \in \minkowski m.$$
This is the $(m+1)$-dimensional {\it Minkowski space}.
Notice that, in contrast to the familiar dot product on Euclidean space, the bilinear form of Minkowski space is not positive definite, i.e. there exist vectors $\asvector v$ such that $\minkowskiform m{\asvector v} {\asvector v} < 0$.
Indeed, for any $\rho > 0$, the {\it $m$-dimensional hyperboloid of radius $\rho$}, denoted $\hyperboloid \rho m$, is a collection of such points:
$$ \hyperboloid \rho m = \{\, \asvector x \in \minkowski m \mid \minkowskiform m {\asvector x} {\asvector x} = - \rho^2,\ x_{m+1} > 0 \,\}. $$
The tangent space to a point $\asvector x \in \hyperboloid \rho m$ is the set of all vectors perpendicular to $\asvector x$ with respect to the bilinear form of the ambient
$$\tangentspace x \hyperboloid \rho m = \{\, {\asvector v} \in \minkowski m \mid \minkowskiform m {\asvector x} {\asvector v} = 0 \,\}.$$
Each tangent space inherits the bilinear form from the ambient, and this restriction is positive-definite (thus $\hyperboloid \rho m$ is a Riemannian manifold, embedded in the {\it pseudo}-Riemannian ambient $\minkowski m$).
In particular, for any $\asvector x \in \hyperboloid \rho m$, it makes sense to talk about the norm $\| \asvector v \|$ of a tangent vector $\asvector v \in \tangentspace {\asvector x} \hyperboloid \rho m$.
Moreover, the distance between two points is given by
\begin{equation}\label{hyperboloiddistance}
	\dist_{\hyperboloid \rho m} (\asvector x, \asvector y) = \rho \arccosh(- \rho^{-2} \minkowskiform m {\asvector x} {\asvector y}), \quad \asvector x, \asvector y \in \hyperboloid \rho m.
\end{equation}

The exponential map is used to move away from a point in the direction of a tangent, and is central to our gradient-based optimisation.
For a point $\asvector x\in \hyperboloid \rho m$, the exponential map
$$\Exp_x : \tangentspace x {\hyperboloid \rho m} \rightarrow {\hyperboloid \rho m}$$
is given by
\begin{equation}\label{Equation:exponential}
\Exp_{\asvector x} (v) = \cosh (\rho^{-1} \| \asvector v \|) \asvector x + \rho \sinh (\rho^{-1} \| \asvector v \|) \frac{\asvector v}{\| \asvector v \|}.
\end{equation}
for $v \ne 0$, and $\Exp_x (0) = x$.
The exponential $\Exp_{\asvector x} (v)$ computes the point that is at distance $\| \asvector v \|$ from $x$ along the geodesic passing through $x$ in the direction of $\frac{\asvector v}{\| \asvector v \|}$.
The inverse of the exponential map $\Exp_x$ is given by the logarithm $\Log_x : \hyperboloid \rho m \rightarrow \tangentspace x {\hyperboloid \rho m}$, given by
\begin{equation}\label{Equation:logarithm}
\Log_x (y) = \frac{\dist (x, y)}{\sqrt{(\rho^{-1}{\minkowskiform m x y})^2 - \rho^2}} (y - \rho^{-2} {\minkowskiform m x y} x)
\end{equation}
for any $y \ne x$, and $\Log_x (x) = 0$.  Finally, since the logarithm and exponential are inverse, it follows that 
\begin{equation}\label{Equation:logarithm-norm}
\| \Log_x (y) \| = \dist (x, y), \quad x, y \in \hyperboloid \rho m.
\end{equation}

\subsection*{Curvature and radius}
The (sectional) curvature of a Euclidean sphere is positive and has an inverse relationship to its radius.  As the radius grows, the curvature approaches zero (i.e. the sphere resembles more and more a flat Euclidean space), On the other hand, if the radius is allowed to shrink towards zero, the curvature will increase without bound.
The role of the parameter $\rho$ in the definition of $\hyperboloid \rho m$ is analogous to that of the radius of the sphere.  Indeed, while the curvature of the hyperboloid is always {\it negative}, it approaches zero as $\rho$ grows, and becomes greater in magnitude (i.e. more negative) as $\rho$ approaches zero.
Indeed, it can be shown that the sectional curvature of $\hyperboloid \rho m$ is  $-\rho^{-2}$.
The significance of $\rho$ in relation to the 4PC is discussed in Section \ref{Section:four-point-condition}.

\subsection*{Relationship of the Poincaré ball and the hyperboloid}
The Poincaré ball $\pb \rho m$ and hyperboloid $\hyperboloid \rho m$ model the same geometry.
Points on the hyperboloid map to points on the Poincaré ball according to the formula:
\begin{equation}\label{hyperboloidtopb}
	\hyperboloid \rho m \to \pb \rho m, \quad \asvector x \mapsto \frac{\rho}{x_{m+1} + \rho}(x_1, \ldots, x_m), \quad \asvector x \in \hyperboloid \rho m,
\end{equation}
while the inverse of this map is given by:
\begin{equation}\label{pbtohyperboloid}
	\pb \rho m \to \hyperboloid \rho m, \quad \asvector y \mapsto \frac{2\rho}{1 - \rho^{-2} r^2}\left(y_1, \ldots, y_m, \frac{1 + \rho^{-2} r^2}{2}\right), \quad \asvector y \in \pb \rho m,
\end{equation}
where $r = \| \asvector y \|$ is the Euclidean norm of $\asvector y$. 

\section*{Appendix: Proofs}
\subsection*{Proof of Theorem \ref{Theorem:hyperbolicity-hyperbolic}}
\begin{proof}
We first consider the case where $\rho=1$.
$\hyperboloid 1 m$ is $\delta$-hyperbolic for some $\delta$ if and only if all triangles in the
space are $\delta'$-slim\cite{Alonsoetal}, for some $\delta'$.
Since the edges of a triangle determine a geodesic subspace of local dimension 2, it suffices to consider the case of $\hyperboloid 1 2$.
This case is well-known exercise on the hyperbolic plane,
treated for example in \cite{Loustau2020hyperbolic}, Exercise 11.6.
Thus, for any $m \geqslant 2$, $\hyperboloid 1 m$ is $\delta$-hyperbolic for some $\delta$.

For the case of arbitrary $\rho$, notice that for $x \in \minkowski m$, by
definition of the hyperboloid, $x \in \hyperboloid 1 m$ if and only if $\rho
x \in \hyperboloid \rho m$.
Moreover, if $x, y \in \hyperboloid 1 m$, then by equation \eqref{hyperboloiddistance},
$$
\dist_{\hyperboloid 1 m} (x, y) = \rho^{-1} \dist_{\hyperboloid \rho m} (\rho x, \rho y).
$$
Thus $(\pointi)$ is a point configuration in $\hyperboloid 1 m$ with pairwise
distance matrix $D$ if and only if $(\rho \pointi)$ is a point configuration
in $\hyperboloid \rho m$ with pairwise distance matrix $\rho D$.
It follows that $\hyperboloid 1 m$ is $\delta$-hyperbolic if
and only if $\hyperboloid \rho m$ is $\rho \delta$-hyperbolic.
This completes the proof of both claims.
\end{proof}

\subsection*{Proof of Proposition \ref{Proposition:gradient-formula}}
	\begin{proof}
	For part \ref{generalgradient}, recall the matrix equation
	$\frac{\partial}{\partial t} \probamatrix (t) = Q\probamatrix (t)$.  For any bases $a,b$,
	then, $\frac{\partial}{\partial t} \proba {ab} (t) = (Q\probamatrix (t))_{ab}$ and so
	$\log \proba {ab} (t) = \frac{(Q\probamatrix(t))_{ab}}{\proba {ab}(t)}$ by the chain rule.
	The claim then follows since the $Z_{ij}$ are constants with respect
	to the points of the configuration $x$.
	Part \ref{jc69gradient} concerns the Jukes-Cantor model, where we have
	\begin{equation}
	  (Q\probamatrix)_{ab}(t) =
	  \begin{cases}
	    - e^{-4t/3}, & \text{if}\ a=b, \\
	    \frac{1}{3} e^{-4t/3}, & \text{otherwise.}
	  \end{cases}
	\end{equation}
	Thus
	\begin{equation*}
	\begin{split}
	\sum_{\sitessigma} \frac{(Q\probamatrix (d_{ij}))_{\sigmai \sigmaj}}{\proba {} (d_{ij})_{\sigmai \sigmaj}} & =
	\sum_{\sigma : \sigmai \neq \sigmaj} \frac{e^{-4t/3}}{3 \probaoffdiag (d_{ij})} - 
	\sum_{\sigma : \sigmai = \sigmaj} \frac{e^{-4t/3}}{\probadiag (d_{ij})} \\
	& = L e^{-4t/3} \left( \frac{r_{ij}}{3 \probaoffdiag (d_{ij})} - \frac{1 - r_{ij}}{\probadiag (d_{ij})}\right).
	\end{split}
	\end{equation*}
	The claim then follows by the first part.
	\end{proof}

\subsection*{Proof of Proposition \ref{Proposition:distance-gradient}}
\begin{proof}
The magnitude of the gradient vector gives the rate of change in that direction with respect to unit change in distance.
Hence $\| \gradient x {\distfnsinglevar y} \| = 1$.
Furthermore, $\distfnsinglevar y$ measures distance along geodesics passing through $y$, and the geodesic connecting $x$ to $y$ has tangent $\Log_x (y)$ at $x$.
Hence $\gradient x {\distfnsinglevar y}$ and $- \Log_x y$ point in the same direction.  Thus
$ \gradient x {\distfnsinglevar y} = \frac{- \Log_x y}{\| \Log_x y \|}, $
and the result follows from equations \eqref{Equation:logarithm} and \eqref{Equation:logarithm-norm}.
\end{proof}

\end{document}